\pdfoutput=1

\documentclass[11pt]{article}
\usepackage[final]{acl}
\usepackage{tabularx}
\usepackage{times}
\usepackage{latexsym}
\usepackage[T1]{fontenc}
\usepackage{float}
\usepackage{svg}
\usepackage[utf8]{inputenc}
\usepackage{microtype}
\usepackage{inconsolata}
\usepackage{graphicx}
\usepackage{quoting}
\usepackage{booktabs}
\usepackage{multirow}
\usepackage{makecell}
\newcommand\z{\allowbreak\hspace{0pt}}
\usepackage{nicematrix}
\usepackage{tabularray}
\usepackage{longtable}
\usepackage{pifont}
\usepackage{tikz}

\newcommand{\cmark}{\ding{51}}%
\newcommand{\xmark}{\ding{55}}%
\newcommand{\tr}[1]{\textcolor{red}{#1}}

\usepackage{tcolorbox}
\usepackage{listings}
\usepackage{xcolor}
\usepackage{fancyvrb}
\definecolor{lightblue}{RGB}{46,110,187}
\definecolor{darkred}{RGB}{150,38,31}
\definecolor{darkgreen}{rgb}{0.0, 0.75, 0.0}
\definecolor{blue}{HTML}{3572EF}
\newcommand{\tg}[1]{\textcolor{darkgreen}{#1}}
\newcommand{\tb}[1]{\textcolor{blue}{#1}}
\usepackage{lipsum}

\title{ConMeC: A Dataset for Metonymy Resolution with Common Nouns}

\author{Saptarshi Ghosh \and Tianyu Jiang \\
  University of Cincinnati \\
  \texttt{ghosh2si@mail.uc.edu, tianyu.jiang@uc.edu} \\}

\begin{document}
\maketitle
\begin{abstract}

Metonymy plays an important role in our daily communication. People naturally think about things using their most salient properties or commonly related concepts. For example, by saying ``The \textit{bus} decided to skip our stop today,'' we actually mean that the bus driver made the decision, not the bus. Prior work on metonymy resolution has mainly focused on named entities. However, metonymy involving common nouns (such as \textit{desk}, \textit{baby}, and \textit{school}) is also a frequent and challenging phenomenon. We argue that NLP systems should be capable of identifying the metonymic use of common nouns in context. We create a new metonymy dataset ConMeC,\footnote{\url{https://github.com/SaptGhosh/ConMeC}} which consists of 6,000 sentences, where each sentence is paired with a target common noun and annotated by humans to indicate whether that common noun is used metonymically or not in that context. We also introduce a chain-of-thought based prompting method for detecting metonymy using large language models (LLMs). We evaluate our LLM-based pipeline, as well as a supervised BERT model on our dataset and three other metonymy datasets. Our experimental results demonstrate that LLMs could achieve performance comparable to the supervised BERT model on well-defined metonymy categories, while still struggling with instances requiring nuanced semantic understanding.

\end{abstract}

\section{Introduction}

Metonymy is often described as one type of figure of speech that is everywhere in our language. As pointed out by~\citet{lakoff1980}, metonymy, like metaphor, is part of our everyday way of thinking and is grounded in our experience~\citep{radden}. People use metonymy when they refer to one concept or word with something else in a close semantic relation. For example, consider the following sentences:

\begin{quoting}[font={itshape, raggedright, noindent}, leftmargin=1em, rightmargin=0in]
(a) The \textbf{desk} called the school police when the babies got into a fight.
\vspace{0.5em}

(b) The \textbf{magazine} wrote, ``She turned thousands of teens into readers.''
\vspace{0.5em}

(c) This \textbf{dish} may be prepared with salmon, stingray and even puffer fish.
\vspace{0.5em}
\end{quoting}

In sentence (a), ``\textit{desk}'' refers to people working at the front desk. In sentence (b), ``\textit{magazine}'' refers to magazine writer. In (c), ``\textit{dish}'' represents the meal being prepared. People don't eat the ``dish'' itself, but rather the meal served in it. 

Previous work has demonstrated that understanding metonymy is crucial for many NLP tasks, such as information extraction~\citep{leveling2008metonymy} and named entity recognition~\citep{gritta2018s}. Metonymy resolution is often formatted as a classification task: given one sentence and a target word in it, the system should identify whether the target word is used in a metonymic way.

Existing datasets for metonymy resolution focus on named entity target word, especially location names, such as SemEval 2007 Shared Task 8~\citep{markert-nissim-2007-semeval}, RelocaR~\citep{gritta-etal-2017-vancouver}, and WIMCOR~\citep{alex-mathews-strube-2020-large}; or are small and composed of handcrafted simple sentences~\citep{pedinotti-lenci-2020-dont}. These limitations pose challenges for developing robust machine learning models that can handle a wide range of metonymy types and their diverse manifestations. The narrow focus also makes it difficult to evaluate models' true capabilities in resolving metonymy in natural language.

To address this issue, we introduce a new dataset that focuses on metonymy resolution with a common noun target. Our dataset consists of 6,000 sentences from Wikipedia, where each sentence is paired with a target common noun and annotated by human annotators to indicate whether that common noun is used metonymically or not. It is not an easy task to extract diverse metonymic sentences so we propose an approach to use large language models (LLMs) with data augmentation to retrieve potential metonymic sentences from a large corpus. Inspired by recent advances in larger language models, we introduce a chain-of-thought based prompting pipeline to detect metonymy. Specifically, we prompt the model to identify the target word's semantic category and then apply the corresponding chain-of-thought prompt. We apply a self-consistency with the majority vote strategy to improve our model's performance. We evaluate the proposed LLM-based method and a supervised transformer-based model on multiple metonymic datasets (including ours). The experimental results show that LLMs can achieve comparable results to fine-tuned transformer models when the target word belongs to specific semantic categories such as container, location and product. It is still challenging for LLMs to grasp the nuanced semantic subtleties. We also show that when the transformer-based model is fine-tuned in a cross-category setup, it experiences a significant performance drop, performing worse than LLM-based methods.

In summary, our contributions are three-fold:

\begin{enumerate}
    \item We introduce a dataset of 6,000 sentences with human annotation indicating whether the target word is metonymic. The dataset is freely available at: \url{https://github.com/SaptGhosh/ConMeC}.

    \item We propose using large language models with a chain-of-thought based prompting method and a self-consistency strategy for this task.

    \item We conduct extensive experiments on our dataset, as well as three other metonymy datasets. Our analysis provides insightful findings for future research on the task.
\end{enumerate}

\section{Related Work}
Metonymy has long been recognized as a significant concept across multiple disciplines,  including philosophy, linguistics, and psychology~\citep{noppen1985metaphor}. Unlike metaphor, metonymy is often approached as a cognitive and pragmatic phenomenon, rather than purely linguistic terms~\citep{maria2015metonymy}. For example, \citet{radden} describes metonymy as a conceptual phenomenon and cognitive process.  \citet{papafragou1996metonymy} defines metonymic use as introducing a new name – like a nickname – for the entity the speaker has in mind. 

Early in the field of artificial intelligence and natural language processing (NLP), multiple attempts have been made to address the ambiguity issues caused by metonymy, such as collative semantics~\citep{fass-1991-met} and local pragmatics~\citep{hobbs1987local}. More recent work in the NLP community follows \citet{markert-nissim-2002-metonymy} that treats the metonymy resolution as a classification task and focuses on metonymies triggered by named entities. SemEval 2007 Shared Task 8~\citep{markert-nissim-2007-semeval} created a metonymy dataset with sentences extracted from British National Corpus (BNC) using keywords of country and company
names. Later, more work has been done in constructing metonymic datasets, including RelocaR \citep{gritta-etal-2017-vancouver} and \textsc{wimcor}~\citep{alex-mathews-strube-2020-large}, both of which focuses primarily on named entities referring to geographical locations or organizations. A number of methods have been proposed to tackle the task, including using syntactic roles and morphological features~\citep{nicolae-etal-2007-utd}, external knowledge~\citep{nastase-strube-2009-combining}, word embeddings~\citep{gritta-etal-2017-vancouver}, as well as transformer-based model~\citep{li-etal-2020-target}.

Our work is most closely related to \citet{pedinotti-lenci-2020-dont}. They created a dataset of 509 artificially constructed metonymic and literal sentence pairs, and proposed a method to determine metonymy by using the contextualized word embeddings. They computed the cosine similarity between the metonymic word and actual paraphrased meaning. For example, in sentence ``\textit{the man sips the \textbf{glass}}'', they replace ``\textit{glass}'' with ``\textit{wine}'', and assume that ``\textit{glass}'' should be more semantically similar to ``\textit{wine}'' in this sentence than a literal sentence like ``\textit{the man grabbed the \textbf{glass}}''. This strategy requires a corresponding literal sentence for each metonymic instance and does not generalize effectively to complex real-world text. In comparison, our dataset builds upon their work by first extracting verbs and nouns from their dataset, using LLMs to augment these pairs, and then using the expanded set as keywords to extract naturally occurring sentences from Wikipedia, which encompasses more complicated and nuanced instances of metonymy. We then used a chain-of-thought prompting technique~\citep{wei-chain-of-thought-prompting} 
to build a 2-step pipeline for identifying metonymy in these sentences.

\section{Dataset Creation}
In this work, we created a new metonymic resolution dataset,~\textbf{ConMeC}~(\textbf{Co}mmon \textbf{N}ouns \textbf{Me}tonymy \textbf{C}orpus), with gold standard human annotations (Metonymic vs. Non-Metonymic) for 6,000 sentences extracted from Wikipedia. Table~\ref{tab:ex_anno} shows some examples of our gold dataset. As far as we are concerned, it is the largest metonymic resolution dataset that focuses on common nouns compared to previous work on named entity metonymy. We believe that the dataset would not only be used as a benchmark for intrinsic evaluation on metonymic resolution, but also serve as a valuable resource to assess NLP pipelines' ability to understand implicit languages in a human-like manner.

\subsection{Metonymy Types}
We followed \citet{pedinotti-lenci-2020-dont} to focus on the six most common types of metonymy: \textsc{container}-\z\textsc{for}-\z\textsc{content}, \textsc{producer}-\z\textsc{for}-\z\textsc{product}, \textsc{product}-\z\textsc{for}-\z\textsc{producer}, \textsc{location}-\z\textsc{for}-\z\textsc{located}, \textsc{causer}-\z\textsc{for}-\z\textsc{result}, and \textsc{possessed}-\z\textsc{for}-\z\textsc{possessor}. For example, ``\textit{the man sips the glass}'' belongs to the \textsc{container-\z for-\z content} category because ``glass'' actually refers to the drink contained in the glass. In the sentence ``\textit{the singer told the magazine}'', the ``magazine'' refers to the producer of the magazine, i.e., the journalist, so it is categorized as \textsc{product-\z for-\z producer}. For simplicity, in the rest of the paper, we will use the first word of the type name interchangeably (such as \textsc{container}).

\subsection{Data Augmentation Using LLM}
It is not an easy task to build a dataset of non-artificial metonymic sentences due to their relative scarcity in typical discourse. To begin with, we collected the common nouns and the associated activity verbs from the handcrafted metonymic sentences in \citet{pedinotti-lenci-2020-dont}, such as $<$glass, sip$>$ and $<$magazine, write$>$. Though they do not always guarantee metonymy, we use them as effective seeds for identifying potential metonymic instances with a higher likelihood. Their dataset contains 509 artificial examples of metonymy with 221 common noun entities associated with 354 verbs, creating 368 unique noun-verb pairs. We can then collect the sentences containing these $<$noun, verb$>$ pairs from a large corpus. However, one limitation of this extraction method is the lack of expression diversity. To include more diverse metonymic usage, we applied a data augmentation technique to increase noun-verb combinations for extraction.

Our goal is to take advantage of large language model's generalization ability to learn more metonymic events. Specifically, we used Llama-3.1-8B~\citep{llama3.1} to generate three alternate words in place of the noun in the existing metonymic sentence from the \citet{pedinotti-lenci-2020-dont} dataset. We then fed the new sentences into Llama to generate three alternative verbs. For example, suppose the original metonymic sentence is  ``\textit{the man sips the glass}''. We would first learn replacement nouns such as \textit{mug, cup} and \textit{tankard}. Then, by asking Llama to replace the verb in the new sentences with augmented nouns, we can retrieve more activity verbs associated with the nouns, such as \textit{taste, drink} and \textit{gulp}. The augmentation pipeline generates a total of 611 nouns, 1,109 verbs, and 4,397 unique noun-verb pairs. 

\begin{figure}
    \centering
    \includegraphics[width=0.85\linewidth]{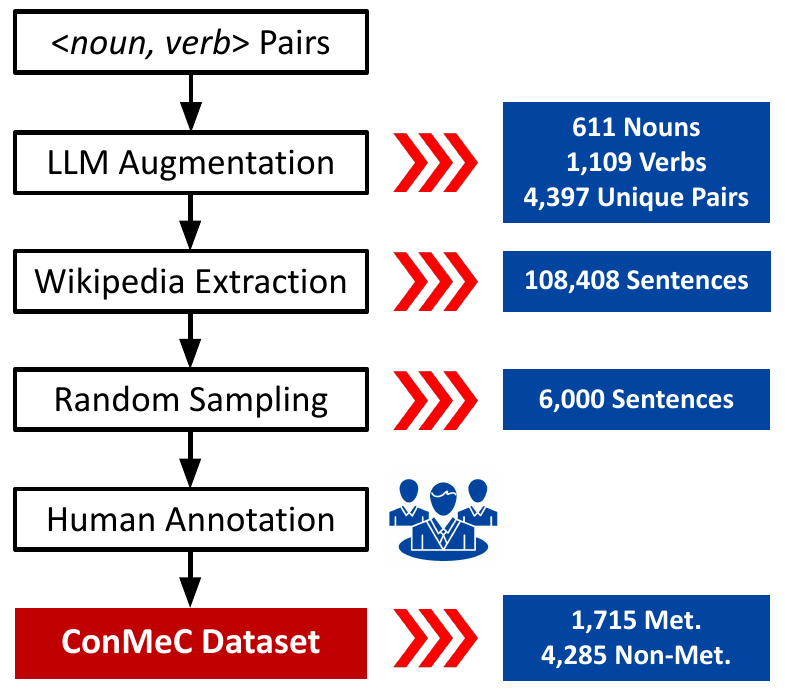}
    \caption{Process of dataset creation.}
    \label{fig:dataset_creation}
\end{figure}

We then used the spaCy~\citep{spacy2020} dependency parser to extract all sentences with these noun and verb pairs from Wikipedia (dump as of 2022-03-01). We required that there exists a dependency relation between the noun and verb in the sentence. This generated a total of 108,408 sentences, with 418 unique nouns, 564 unique verbs, and 2,240 unique $<$noun, verb$>$ pairs over six metonymy types. For the annotation purpose, we randomly sampled 1,000 sentences from each metonymic category (6,000 sentences in total). To avoid the dominance of some commonly used nouns such as ``\textit{dish}'' and ``\textit{city}'', our sampling method adopted a uniform distribution across different nouns. Our data collection pipeline is shown in Figure~\ref{fig:dataset_creation}.

\begingroup

\begin{table*}[t]
    \NiceMatrixOptions {
    custom-line = {
       command = dashedmidrule ,
       tikz = { dashed } ,
       total-width = \pgflinewidth + \aboverulesep + \belowrulesep ,
     } }
    \centering
    \small
    \resizebox{\textwidth}{!}{
    \begin{NiceTabular}{p{0.85\linewidth}|c|l}
    \toprule
    \multicolumn{1}{c|}{\bf Sentence} & \multicolumn{1}{c|}{\bf Met.} &
    \multicolumn{1}{c}{\bf Met. Type} \\
    \midrule
    (1) This \tb{\bf{dish}} tastes best when it is dipped in a mixture of soy sauce, vinegar, and red pepper powder. & \tg{\cmark} & \textsc{container} \\
    (2) Immediately after the delivery of the baby, the baba fills a clay \tb{\bf{jug}} with water, dips a sprig of basil or geranium in it and takes it to the church. & \tr{\xmark} & \\ 
    \dashedmidrule
    (3) While there, he read the socialist \tb{\bf{historian}} Guglielmo Ferrero in depth. & \tg{\cmark} & \textsc{producer} \\
    (4) The late \tb{\bf{musician}}, filmmaker, and photographer Jon Sholle praised the ``vision'' and ``concept'' of Karp's photographs. & \tr{\xmark} & \\
    \dashedmidrule
    (5) The \tb{\bf{media}} ridiculed the case, and she appeared several times on late-night television. & \tg{\cmark} & \textsc{product} \\
    (6) Griffin also engaged in Holocaust denial, publishing articles promoting such ideas in The Rune, a \tb{\bf{magazine}} produced by the Croydon BNP. & \tr{\xmark} & \\
    \dashedmidrule
    (7) When we played Wembley, Salman showed up in person and the \tb{\bf{stadium}} erupted. & \tg{\cmark} & \textsc{location} \\
    (8) Ezra returns to his \tb{\bf{office}} where he talks with Jackie, and tells her that he is extremely angry with her. & \tr{\xmark} & \\
    \dashedmidrule
    (9) As he began his questioning of the witnesses, the \tb{\bf{crowd}} drowned out his voice and surrounded him. & \tg{\cmark} & \textsc{causer} \\
    (10) The Temple was built from stone made ready at the quarry, and no \tb{\bf{hammer}}, ax, or other iron tool was heard at the building site. & \tr{\xmark} & \\
    \dashedmidrule
    (11) Willis, a former medical student, was working as hired \tb{\bf{muscle}} for Two-Face and had disappeared suspiciously following a botched assignment. & \tg{\cmark} & \textsc{possessed} \\
    (12) The decision was reached as the \tb{\bf{gun}} could achieve up to ten rounds per minute rate of fire. & \tr{\xmark} & \\
    \bottomrule
    
    \end{NiceTabular}
    }
    \caption{A sample of the annotated examples. The target words are marked in \textbf{\tb{blue}}. The second column shows the annotated label for metonymy (\tg{\cmark}) vs. non-metonymy (\tr{\xmark}). If yes, the third column shows the metonymy type.}
    \label{tab:ex_anno}
\end{table*}
\endgroup

\subsection{Human Annotation}
We recruited two human annotators to complete the annotation task. To prepare the annotators, we presented them with the definitions of metonymy and a few examples for each category. Each time, annotators were given one sentence and one target word (noun) in the sentence, and they were asked to determine whether the target word was used metonymically or not. When the annotations were finished, we measured the inter-annotator agreement using Cohen's kappa. Different metonymy types received variant kappa scores, with the highest being \textsc{location-for-located} (0.94) and lowest being \textsc{possessed-for-possessor} (0.75). The overall kappa score is 0.83.

To create the final set of gold standard labels, we had the annotators adjudicate their disagreements. Table~\ref{table:dataset_dist} shows the number of metonymic sentences for each type in our final dataset. Each category consists of 1,000 sentences, with \textsc{location} having the most (454) and \textsc{producer} having the least (129) number of metonymic sentences. In general, around 28\% of the sentences were annotated as metonymic. Table~\ref{tab:ex_anno} shows some annotated examples in different types.

\begingroup

\begin{table}[t]
   \centering
   \resizebox{0.85\linewidth}{!}{
   \begin{tabular}{lcc}
   \toprule
   & Metonymy & Non-Metonymy \\
   \midrule
   \textsc{container} & 226 & 774 \\
   \textsc{producer} & 129 & 871 \\
   \textsc{product} & 406 & 594 \\
   \textsc{location} & 454 & 546 \\
   \textsc{causer} & 317 & 683 \\
   \textsc{possessed} & 183 & 817 \\
   \midrule
   Total & 1,715 & 4,285 \\
   \bottomrule
   \end{tabular}
   }
\caption{Distribution of metonymic and non-metonymic examples across six categories. Each category contains 1,000 sentences (6,000 annotated sentences in total). }
\label{table:dataset_dist}
\end{table}
\endgroup

\section{Methodology}

\begin{figure*}
    \centering
    \includegraphics[width=0.95\linewidth]{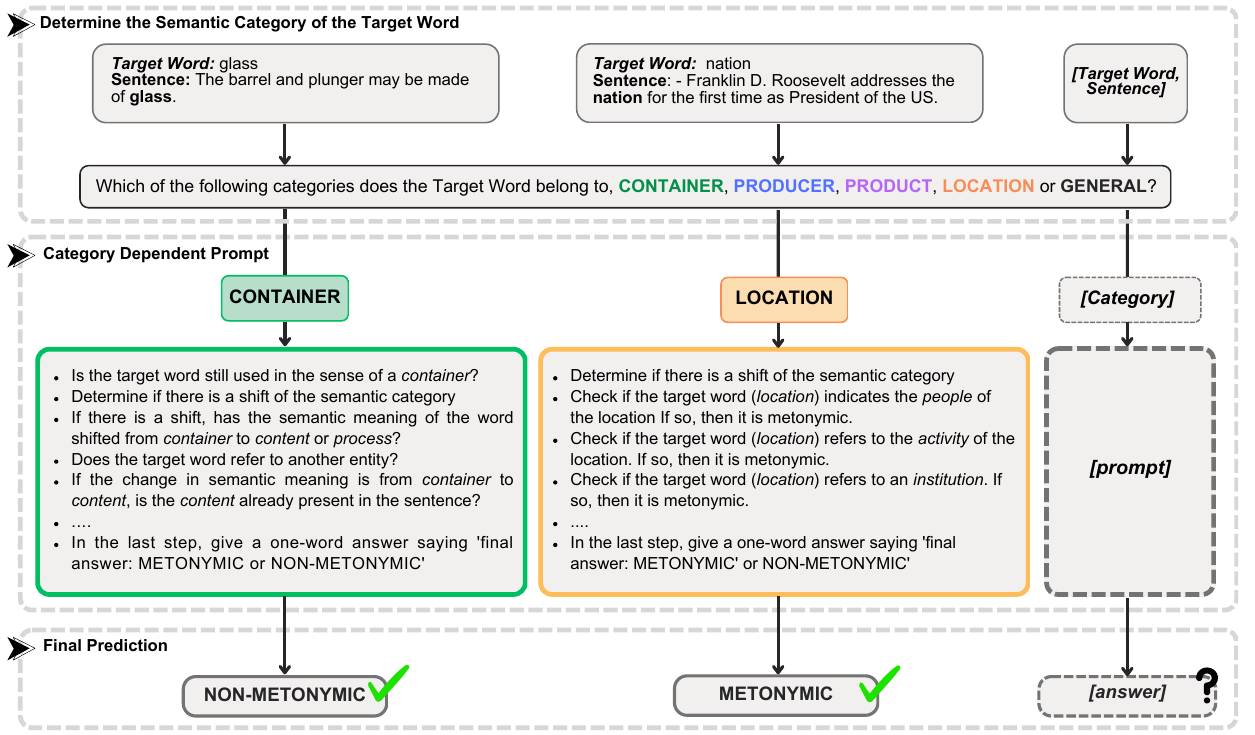}
    \caption{The architecture of our 2-step prompting method. We illustrates two examples. The LLM will first determine the semantic category of the target word in the sentence, such as \textsc{container} or \textsc{location}. Then, given the category dependent prompting, the model should predict whether there exists a metonymy use or not.}
    \label{fig:pipeline_and_prompt}
\end{figure*}

In this work, we focused on investigating how state-of-the-art large language models perform on the task of metonymy resolution. We explored several approaches to prompt the large language model. We first present a method based on the observation that metonymic sentences typically trigger a semantic category change in the target word. We then introduce a 2-step prompting approach, where the model is first asked to categorize the target noun and then detect any semantic category change. We further show that our model's performance can improve through a self-consistency with majority vote approach. Finally, we fine-tune a BERT model on our dataset to compare with large language models approach.

\subsection{Chain-of-Thought Prompting}

We initially explored prompting large language models with the definition of metonymy and a few examples~\citep{brown2020-fewshot}, then asking the model to classify new instances. However, the model struggled with a consistent understanding of the metonymy phenomenon. Inspired by the success of chain-of-thought~\citep{wei-chain-of-thought-prompting} in many NLP tasks, we propose a chain-of-thought based prompting method.

Since the most familiar definition of metonymy is ``using one entity to refer to another that is related to it''~\citep{lakoff1980}, an intuitive idea is to use chain-of-thought to prompt the large language models to identify if there exists a \textit{reference shift}, in which a linguistic sign refers not only to its default concept, but to another concept~\citep{adel2014}. However, a direct instruction about a reference shift will cause the LLMs to generate excessive false positives. For example, given the sentence ``\textit{she browses a \textbf{newspaper}}'', the model will predict yes and explain that it contains a reference shift because the newspaper refers to the content of the newspaper.

The idea of our method is based on the observation that reference shift often co-occurs with a change of semantic categories, which serves as a key indicator for identifying metonymic expressions. This observation is aligned with previous work on the cognitive bases of metonymy describing it as ``a specific semantic relation'' \citep{koch1999frame}.  
For example, in the sentence ``\textit{the \textbf{city} celebrated the festival}'', the target word (noun) ``city'' refers to the people in the city, not city as a location. Similar situation exists for the ``church'' in \textit{``the man devotedly followed the \textbf{church}''}. The shift from one semantic category (location) to another (people/organization) provides a clue for detecting metonymy.

Building on this insight, we designed a prompt to capture these transitions between semantic categories. The prompt guides the model to first identify any shift between semantic categories in the context of the target word through a series of reasoning steps, and then determine if this categorical transition indicates a metonymic usage. 

\begin{table*}
    \centering
    \resizebox{0.8\linewidth}{!}{
    \begin{tabular}{lcccccccc}
    \toprule
    \multirow{3}{*}{\textbf{Model}} & \multirow{3}{*}{\textbf{Acc}} & \multirow{3}{*}{\textbf{Macro-F1}} & \multicolumn{3}{c}{\bf Metonymic} & \multicolumn{3}{c}{\bf Non-Metonymic} \\
    \cmidrule(lr){4-6} \cmidrule(lr){7-9}
    & && Pre & Rec & F1 & Pre & Rec & F1 \\
    \midrule
    Llama-8B-Basic & 61.2 & 61.2 & 39.0 & 60.2 & 47.3 & 79.0 & 71.4 & 75.0 \\
    Llama-8B-CoT & 44.3 & 44.0 & 33.3 & \textbf{91.6} & 48.9 & 87.9 & 25.1 & 39.1 \\
    Llama-8B-CoT-2S & 65.2 & 61.0 & 42.8 & 50.8 & 46.5 & 78.7 & 72.8 & 75.6 \\
    Llama-8B-CoT-2S-SC & 66.5 & 62.0 & 44.3 & 52.8 & 48.1 & 77.6 & 73.6 & 75.9 \\
    \midrule
    Llama-70B-Basic  & 71.5 & 66.5 & 50.7 & 58.2 & 54.2 & 81.8 & 76.3 & 78.9 \\
    Llama-70B-CoT  & 68.8 & 66.6 & 47.6 & 75.2 & 58.3 & 86.7 & 66.1 & 75.0 \\
    Llama-70B-CoT-2S  & 78.1 & 73.6 & 57.5 & 70.6 & 63.3 & 87.9 & 80.2 & 83.9 \\
    Llama-70B-CoT-2S-SC  & 79.4 & 76.5 & 61.9 & 76.3 & 68.3 & 89.1 & 80.6 & 84.6 \\
    \midrule
    GPT-4o & 82.6 & 77.9 & 75.6 & 59.7 & 66.7 & 85.2 & \textbf{92.3} & 88.6 \\
    \midrule
    BERT & \textbf{86.6} & \textbf{82.9} & \textbf{77.4} & 74.2 & \textbf{75.2} & \textbf{89.8} & 91.3 & \textbf{90.6} \\
    \bottomrule
    \end{tabular}
    }
\caption{Results for all models. \textbf{Llama-Basic}: simple prompt asking Llama whether a sentence is metonymic or not. \textbf{Llama-CoT}: one general chain-of-thought (CoT) prompt for all categories. \textbf{Llama-CoT-2S}: 2-step (2S) chain-of-thought prompt. \textbf{Llama-CoT-2S-SC}: 2-step chain-of-thought prompt with self-consistency (SC) method using the majority vote across multiple runs. \textbf{GPT-4o}: 2-step chain-of-thought (CoT-2S) using GPT-4o's API. \textbf{BERT}: supervised BERT with 5-fold cross validation. }
\label{table:main_table}
\end{table*}

\subsection{2-Step Pipeline with Target Word Categorization}

Through semantic based chain-of-thought prompting, we observed that the model was able to identify metonymy more efficiently. However, the model would sometimes miss the nuanced and subtle cases of metonymy due to the complexity of natural language. One common error is caused by word sense disambiguation. For example, in the sentence ``\textit{the \textbf{dish} had to be pointed directly at the satellite, with nothing blocking the signal}'', ``dish'' refers to a satellite dish, not the common dinnerware, and there is no metonymy implied in this instance. However, our model made a wrong prediction by identifying a semantic category change of the word ``dish''. 

To overcome this challenge, we design a 2-step pipeline. We first provide the target word to the LLM and ask it to predict a category from five options: \textsc{container, producer, product, location}, and \textsc{general}. We do not include \textsc{causer} or \textsc{possessed}, as the target words in these categories do not refer to a particular type of semantic category like the others. Based on the selection of the model, we apply a semantic category dependent chain-of-thought prompting at the second stage. We create five different chain-of-thought prompts for each category, instead of one universal chain-of-thought prompt. This way, the model is able to identify the challenging and implicit cases of metonymy more efficiently. Figure \ref{fig:pipeline_and_prompt} shows the architecture of our pipeline with an example of two sentences in different categories.

\subsection{Self-Consistency with Majority Vote}

Previous studies have demonstrated that large language models often suffer from semantic consistency issue, i.e., they can sometimes generate contradictory outputs when presented with prompts with similar or equivalent semantics~\citep{pezeshkpour2023large, yang2024enhancing}. The idea of majority vote and its variants have achieved a good success in improving LLMs performance, such as the self-consistency strategy~\citep{wang2023-chainthought}, multi-agents~\citep{du2023improving}, and self-agreement~\citep{lin-etal-2024-just}. We observed similar inconsistency in our model's output. Following the same idea, we adopted a majority vote method from multiple runs.

\section{Experimental Results}

We evaluate our LLM-based models on our gold labeled data, as well as three other existing datasets for metonymy resolution. We also present results for the fine-tuned BERT as a comparison. For the evaluation metric, we report the precision, recall and F1-score for both metonymic and non-metonymic examples, and an overall accuracy and macro-F1 score.

\subsection{LLMs and Gold Supervision}
Table \ref{table:main_table} shows our experimental results.
The table describes four primary sections: variations of Llama-3.1-8B and Llama-3.1-70B, GPT-4o and the supervised BERT. The first and second sections describe four different setups of Llama-3.1-8B and Llama-3.1-70B respectively: 1) \textbf{Basic}: Simple prompt asking the model if the sentence is metonymic or not; 2) \textbf{CoT}: One general chain-of-thought prompt with semantic category change; 3) \textbf{CoT-2S}: Our 2-step pipeline -- first identify the semantic category of the target word, and then apply the corresponding chain-of-thought prompt; 4) \textbf{CoT-2S-SC}: We run the Llama model multiple times and use the majority vote to determine the final result. We use the temperature 0.4 for the categorization step, and 0.6 for the chain-of-thought step, with top-p 0.9. As can be seen from the table, Llama 70B consistently outperforms Llama 8B. When using the smaller 8B model, the Basic prompt has better performance than CoT. But we get opposite results on the 70B. By adding the 2-step pipeline, Llama-70B-CoT-2S receives 7.0 points higher macro-F1 score than 70B-CoT while Llama-8B barely improves. We believe this is due to the size of the 70B model. Llama-8B and particularly 70B further benefits from the majority vote method.

In the third section, we show the performance of GPT-4o (gpt-4o-2024-05-13) using our 2-step prompting (same as \textbf{CoT-2S}). GPT-4o achieves better performance than Llama, outperforming the best Llama model using majority vote (Llama-70B-CoT-2S-SC) in a single run. Due to budget limitations, we didn't run the self-consistency method with GPT-4o. The breakdown of precision and recall reveals that GPT-4o is more conservative to predict metonymy, leading to higher precision but lower recall for the true metonymy instances compared to Llama. 

The final section shows the performance of supervised method. We fine-tuned BERT-base-uncased~\citep{devlin-etal-2019-bert} with 5-fold cross validation on our dataset. We used random seeds to run the experiments 5 times and report their average score. We used the last hidden state of the \textsc{[cls]} token and appended a linear layer for binary classification.  The learning rate was set at 1e-5 and Adam optimizer was used. Training was conducted for 3 epochs with a batch size of 16. It is not a surprise that supervised BERT has the best performance among all the models, with an overall accuracy of 86.6 and macro-F1 of 82.9.

\begin{table}[t]
    \centering
    \resizebox{0.95\linewidth}{!}{
    \begin{tabular}{lcccc}
    \toprule
    \multirow{2}{*}{\textbf{Dataset}} & \multicolumn{2}{c}{\bf Llama} & \multicolumn{2}{c}{\bf BERT} \\
    \cmidrule(lr){2-3} \cmidrule(lr){4-5}
    & Met & Non & Met & Non \\
    \midrule
    Our dataset (ConMeC) & 63.3 & 83.9 & 75.2 & 90.6 \\
    Pedinotti-Lenci  & 72.7 & 73.6 & 81.9 & 83.4 \\
    ReLocaR & 86.1 & 82.9 & 90.2 & 90.0 \\
    SemEval2007-Task8 & 50.0 & 72.8 & 61.4 & 90.4 \\
    \bottomrule
    \end{tabular}
    }
\caption{F1-score comparison of Llama-70B-CoT-2S and BERT on metonymic and non-metonymic instances across different datasets.
}
\label{table:other_datasets}
\end{table}

\begingroup
\renewcommand\baselinestretch{0.8}
\begin{table*}[ht]
    \NiceMatrixOptions {
    custom-line = {
       command = dashedmidrule ,
       tikz = { dashed } ,
       total-width = \pgflinewidth + \aboverulesep + \belowrulesep ,
     } }
    \centering
    \small
    \resizebox{\textwidth}{!}{
    \begin{NiceTabular}{p{0.85\linewidth}|c|c|c}
    \toprule
    \multicolumn{1}{c|}{\bf Sentence} & \multicolumn{1}{c|}{\bf Gold} &
    \multicolumn{1}{c}{\bf Llama} & \multicolumn{1}{c}{\bf GPT}  \\
    \midrule
    (1) He also enjoyed a \tb{\bf{glass}} of whisky or wine while relaxing. & Non-Met. & \tr{\xmark} & \tg{\cmark} \\ 
    \midrule
    (2) It was the first time a West African \tb{\textbf{artist}} had openly criticized police brutality in popular music. & Non-Met. & \tr{\xmark} & \tg{\cmark} \\ 
    \midrule
    (3) His publications have been cited several ten thousand times, which makes him one of the currently most cited European social \tb{\textbf{scientist}}. & Met. & \tr{\xmark} & \tr{\xmark} \\ 
    \midrule
    (4) She was translated into Danish, Norwegian, German, Russian, French, English, Italian, Dutch, Hungarian and Czech, and was the most widely read Swedish \tb{\textbf{novelist}} of her time. & Met. & \tr{\xmark} & \tr{\xmark} \\
    \midrule
    (5) The novel is set in Sitka, which it depicts as a large, Yiddish-speaking \tb{\textbf{metropolis}}. & Met. & \tr{\xmark} & \tr{\xmark} \\ 
    \midrule
    (6) They will be flown into the stadium with helicopters when a full \tb{\textbf{stadium}} is cheering for them when they arrive. & Met. & \tg{\cmark} & \tr{\xmark} \\ 
    \midrule
    (7) As he began his questioning of the witnesses, the Clodian \tb{\textbf{crowd}} drowned out his voice and surrounded him. & Met. & \tr{\xmark} & \tr{\xmark} \\
    \midrule
    (8) Holland first heard the Native American  \tb{\textbf{flute}} at a concert Webster University near St. Louis in 1994. & Met. & \tr{\xmark} & \tr{\xmark} \\
    \midrule
    (9) He nearly spun out with 20 laps to go but saved the \tb{\textbf{truck}}, later inheriting the lead from Stewart Friesen after he ran out of fuel. & Non-Met. & \tr{\xmark} & \tg{\cmark} \\
    \midrule
    (10) 1942, known as 48 Hours in the USA, in which she is shown wielding a \tb{\textbf{rifle}} to defend a house from German paratroopers. & Non-Met. & \tr{\xmark} & \tg{\cmark} \\
    \bottomrule
    
    \end{NiceTabular}
    }
    \caption{Error analysis of Llama and GPT-4o using the CoT-2S architecture. The target words are marked in \textbf{\tb{blue}}. The second column shows the gold labels. The third and fourth column shows whether the model's prediction is correct (\tg{\cmark}) or incorrect (\tr{\xmark}).}
    \label{tab:error_case}
\end{table*}
\endgroup

\subsection{Evaluation on Other Metonymy Datasets}

We also evaluated the 2-step pipeline model (Llama-70B-CoT-2S) on a few existing metonymy dataset. ReLocaR~\citep{gritta-etal-2017-vancouver} and SemEval 2007 Shared Task 8~\citep{markert-nissim-2007-semeval} are two datasets focusing solely on named entity metonymy, specifically geographical location and organization named entities. Due to this, we modified our 2-step model for these datasets by removing the categorization from the first step, only using the chain-of-thought prompt for the \textsc{location} category, saving significant computational time. ReLocaR has 496 metonymic and 486 non-metonymic examples also extracted from Wikipedia. SemEval contains 167 metonymic and 721 non-metonymic instances from British National Corpus. \citet{pedinotti-lenci-2020-dont} contains 509 metonymic and 509 non-metonymic hand-crafted sentences.

For a comparison, we evaluated the BERT model on each dataset as well. We fine-tuned BERT on each dataset using the same learning rate and optimizer as the previous experiment. For ReLocaR and SemEval2007, we used the original training set given in these datasets to train BERT and reported the result on their test set. For the \citet{pedinotti-lenci-2020-dont} and our dataset, we ran the 5-fold cross validation with random sampling.

Table~\ref{table:other_datasets} shows the F1-score for both metonymic and non-metonymic instances. BERT has superior F1 scores for both metonymic and non-metonymic across all datasets. However, on the ReLocaR dataset, we can see that Llama significantly narrows the gap with the supervised method. Both BERT and Llama receive a lower score on our dataset compared to Pedinotti-Lenci. This is as expected since our metonymic sentences are longer and semantically more complex than artificial sentences such as ``\textit{the man sips the glass}'', making it a more challenging task.

\begin{figure}[t]
    \centering
    \includegraphics[width=0.98\linewidth]{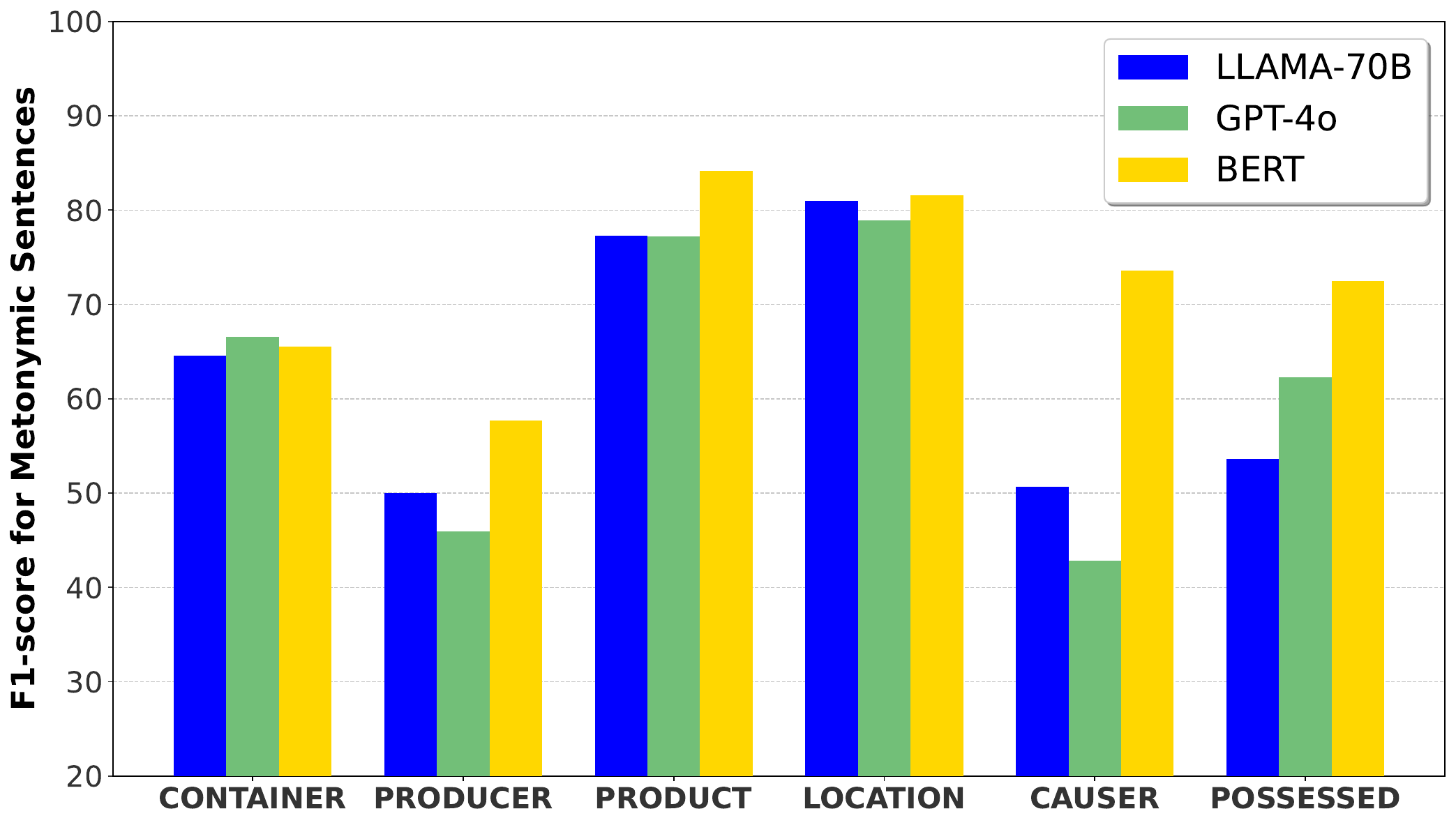}
    \caption{F1-scores across six categories among Llama (blue), GPT-4o (green) and BERT (yellow) on the metonymic sentences.}
    \label{fig:llama_bert_gpt_met}
\end{figure}

\subsection{Performance Breakdown Across All Categories}

Figure~\ref{fig:llama_bert_gpt_met} shows the performance comparison of Llama, GPT-4o and BERT across the six metonymic categories. The Llama scores are reported on the best Llama model (70B-CoT-2S-SC). We can see from the chart that both Llama and GPT-4o achieved comparable results with BERT on the first four categories, with GPT-4o even outperforming BERT on the \textsc{container} category. However, for \textsc{causer} and \textsc{possessed}, LLMs witnessed a big performance drop as compared to BERT. This can be explained by the semantic complexity of these two categories -- \textit{causer} or \textit{possessor} could be objects, people, or many other semantic types. Our 2-step prompt for these two categories does not specify one particular semantic type. Using a general prompt does not cover the complexities associated with metonymy like the four other categories, leading to a low performance. This leaves room for improvement in future work.

\section{Analysis}

We conduct further analysis to better understand the behavior of the models on our dataset. 

\subsection{Error Cases}

Table \ref{tab:error_case} outlines the errors made by Llama-70B and GPT-4o using CoT-2S architecture in a single run. In sentence (1), ``\textit{a glass of whisky}'' should not be considered \textsc{container-for-content} metonymy because the content (\textit{whisky}) is already present in the sentence. Llama still struggles with this type of cases even though our prompt explicitly covers such a scenario. Sentence (2) represents a straightforward example, where Llama made the wrong prediction. From our observation, Llama is more prone to random errors, which can be mitigated by the self-consistency method.

Sentence (3-6) demonstrate that the target word could sometimes be used both metonymically and literally within the same sentence. For instance, ``\textit{Yiddish-speaking metropolis}'' in sentence (5) is a metonymic phrase referring to the people in the metropolis speaking Yiddish, whereas the broader context uses \textit{metropolis} as a location. Similarly in sentences (3) and (4), the metonymy is implied through the phrases ``\textit{most cited European social scientist}'' and ``\textit{most widely read Swedish novelist}''. We further discuss these instances in  Appendix~\ref{app:bothmet}. In sentence (6), the target word ``\textit{stadium}'' appears twice in the sentence - once metonymically and once literally. These scenarios are tricky for models to identify the correct instance of the target word. 

Sentence (7) and (8) are examples of \textsc{causer-for-result} metonymy where most subtleties lie. For example, ``\textit{Holland first heard the Native American flute at a concert...}'', the word \textit{flute} refers to the music caused by playing the instrument, not the instrument itself. The models struggle with these nuanced implication of metonymy. Sentence (9) and (10) belongs to the \textsc{possessed} category, where Llama in particular struggles, and tend to label sentences as metonymic.

\subsection{Cross Category Gold Supervision}

In our experiments with BERT in Table \ref{table:main_table}, we trained the model using 5-fold cross-validation on the entire dataset. The model can train and test on the same metonymic category. To assess BERT's performance in a cross-domain setting, we trained the model on five categories and tested it on the sixth. Table \ref{table:cross_domain_1} shows the F1-score for metonymic sentences under this setup. Comparing the cross category with the general supervised method results, the F1-score experiences a significant drop. This illustrates the limitations of the fine-tuned BERT model, namely its reliance on high quality training data of the same category or type. In contrast, LLMs do not have this constraint, and they also demonstrate a better generalization ability. This flexibility is a significant advantage of LLMs, allowing them to generalize across diverse categories.

\begin{table}[t]
    \centering
    \resizebox{0.99\linewidth}{!}{
    \begin{tabular}{lcccccc}
    \toprule
    \multirow{3}{*}{\textbf{Category}} & \multicolumn{3}{c}{\bf Cross Category} & \multicolumn{3}{c}{\bf General} \\
    \cmidrule(lr){2-4} \cmidrule(lr){5-7}
    & Pre & Rec & F1 & Pre & Rec & F1 \\
    \midrule
    \textsc{container} & 86.5 & 17.4 & 28.3 & 85.7 & 53.1 & 65.5 \\
    \textsc{producer}  & 21.7 & 43.7 & 29.6 & 86.1 & 43.4 & 57.7 \\
    \textsc{product} & 64.5 & 72.7 & 68.2 & 89.9 & 79.3 & 84.2 \\
    \textsc{location} & 84.1 & 47.6 & 60.5 & 89.0 & 75.7 & 81.6 \\
    \textsc{causer} & 51.2 & 32.6 & 40.6 & 78.7 & 69.0 & 73.6 \\
    \textsc{possessed} & 47.4 & 33.5 & 39.7 & 82.0 & 65.0 & 72.5 \\
    \bottomrule
    \end{tabular}
    }
\caption{BERT performance on our dataset. Cross Category: train on 5 categories and test on the rest one category. General: 5-fold cross validation on the whole annotated dataset.}
\label{table:cross_domain_1}
\end{table}

\subsection{Does Context Help?}

\begin{table}[t]
    \centering
    \resizebox{0.99\linewidth}{!}{
    \begin{tabular}{lcccccc}
    \toprule
    \multirow{3}{*}{\textbf{Model}} & \multicolumn{3}{c}{\bf Llama} & \multicolumn{3}{c}{\bf BERT} \\
    \cmidrule(lr){2-4} \cmidrule(lr){5-7}
    & Acc & Met & Non & Acc & Met & Non \\
    \midrule
    With Context & 74.5 & 58.1 & 80.3 & 86.2 & 75.0 & 90.5 \\
    Without Context & 78.0 & 63.3 & 83.9 & 88.6 & 75.2 & 90.6 \\
    \bottomrule
    \end{tabular}
    }
\caption{Overall accuracy and F1-scores of metonymic and non-metonymic sentences for two models with and without the context.
}
\label{table:context_table}
\end{table}

In all previous experiments, we provided the model only with one sentence and its target word to determine metonymy. To assess whether additional context can help the model make predictions, we provide one preceding and one following sentence to the model. Table \ref{table:context_table} shows the results of this context experiment, comparing the performance of our best model (Llama-70B-CoT-2S) and fine-tuned BERT. By adding the context, BERT shows a minor drop in accuracy and F1-score, while the drop in performance for Llama is more noticeable. This indicates that additional context might introduce noise or disrupt LLMs ability to make the correct decision.

\subsection{Number of Votes}

For the self-consistency method, we ran the experiments multiple times for the CoT-2S model to decide on the majority vote. Table \ref{table:ensemble_basic_cot} shows the results of running the self-consistency method on the Llama 70B-Basic, 8B-CoT-2S and 70B CoT-2S models. The performance does not significantly improve for the 70B-Basic and 8B-CoT-2S models. But the 70B-CoT-2S model benefits significantly from the voting, gaining a F1-score of 5.0 on metonymic examples. This is because the chain-of-thought reasoning process can explore multiple reasoning paths when determining metonymy in a sentence. However, the model occasionally follows an incorrect path or logic. Voting helps mitigates this issue. The smaller model 8B-CoT-2S does not exhibit similar improvements on majority voting compared to the 70B-CoT-2S. This observation aligns with previous findings that indicate larger model size generally leads to better performance \citep{scaling_law_paper, scaling_fine_tuned_paper}. 

\section{Conclusion}

We introduced a new dataset, ConMeC, containing 6,000 sentences with human labels for the task of common noun metonymy resolution. We proposed a chain-of-thought inspired prompting method with the state-of-the-art large language models to detect metonymy. We evaluated large language models and a fine-tuned BERT model our dataset, as well as three other metonymy datasets. Our experimental results demonstrate that with careful design, large language models can achieve comparable results as supervised BERT model on particular metonymy categories, although there is still room for improvement that we hope will inspire future work on this task.

\begin{table}[t]
    \centering
    \resizebox{0.98\linewidth}{!}{
    \begin{tabular}{lccc}
    \toprule
    \makecell{No. of \\ Votes} & \makecell{Llama-70B- \\ Basic} & \makecell{Llama-8B- \\ CoT-2S} & \makecell{Llama-70B- \\ CoT-2S} \\
    \midrule
    1  & 54.2 \scriptsize{\hspace{2.5em}}  & 47.3 \scriptsize{\hspace{2.5em}} & 63.3 \scriptsize{\hspace{2.5em}} \\
    3  & 54.3  \scriptsize{(+0.1)} & 48.3  \scriptsize{(+1.0)} & 64.4 \scriptsize{(+1.1)} \\
    5  & 54.4  \scriptsize{(+0.2)} & 47.9  \scriptsize{(+0.6)} & 65.1 \scriptsize{(+1.8)} \\
    7  & 54.5  \scriptsize{(+0.3)} & 48.0  \scriptsize{(+0.7)} & 67.4 \scriptsize{(+4.1)} \\
    9  & 54.4  \scriptsize{(+0.2)} & 48.1  \scriptsize{(+0.8)} & 68.3 \scriptsize{(+5.0)} \\
    \bottomrule
    \end{tabular}
    }
\caption{Increase in F1-score on metonymic examples using majority vote. The values in parentheses indicate the increase in F1-score for metonymic examples compared to the single-run baseline.}
\label{table:ensemble_basic_cot}
\end{table}

\section*{Limitations}

The use of prompting with large language models requires substantial computational time. In our best model setting (Llama-70B-CoT-2S), it takes around 40 hours to inference on 6,000 examples with 2 NVIDIA A100 GPUs. This time requirement presents challenges for analyzing extensive datasets or extracting metonymic sentences from large-scale corpora.

In this work, we have restricted our investigation to focus on six common categories of metonymy, even though our proposed method can be potentially be adapted to other categories. We would like to extend our work to study a diverse range of metonymic expressions. We will leave this for future work.

We have not yet evaluated the effectiveness of our model on any downstream NLP applications. It remains a curious question as how the integration of our metonymy resolution system could affect the overall performance on existing NLP benchmarks. Metonymy is still a challenging problem for NLP. We hope our work can attract more attention to this problem.

\section*{Acknowledgments}
We thank the Cincinnati NLP group for their constructive comments. We also thank the anonymous NAACL
reviewers for their valuable suggestions and feedback.

\bibliography{naacl}

\clearpage
\newpage  
\appendix

\section{Prompts for Our Best Model}

Table \ref{tab:all_prompts} features all the prompts in our models. In CoT model, only the \textsc{general} prompt is used, while all categories are used in CoT-2S model.

For negative examples, we instruct the model to provide a one-word answer in the last step, saying the sentence is ``\textsc{literal}'', instead of ``\textsc{non-metonymic}'', as shown in Figure \ref{fig:pipeline_and_prompt}. This ensures that word matching can detect the model's final prediction, even if it does not follow the instruction to give a one-word response in the last step.

\section{DeepSeek Comparison}

With the recent launch of DeepSeek R1 in January 2025 \cite{deepseekai2025}, the model has demonstrated performance on par with GPT-4o across a wide range of tasks, including math solving, code generation, and natural language understanding. Given its promising capabilities, we evaluate its effectiveness on our dataset to compare its performance against Llama and GPT-4o.

Table \ref{table:deepseek_table} presents the results of a single-run evaluation of DeepSeek R1 on our dataset, compared with Llama and GPT-4o. For this experiment, we utilized our best performing 2-step chain-of-thought architecture with the DeepSeek 70B distilled model.

\begin{table}[ht]
    \centering
    \resizebox{0.99\linewidth}{!}{
    \begin{tabular}{lcccccc}
    \toprule
    \multirow{3}{*}{\textbf{Model}} & \multicolumn{3}{c}{\bf Metonymic} & \multicolumn{3}{c}{\bf Non-Metonymic} \\
    \cmidrule(lr){2-4} \cmidrule(lr){5-7}
    & Pre & Rec & F1 & Pre & Rec & F1 \\
    \midrule
    Llama & 57.5 & \textbf{70.6} & 63.3 & \textbf{87.9} & 80.2 & 83.9 \\
    GPT-4o & \textbf{75.6} & 59.7 & \textbf{66.7} & 85.2 & \textbf{92.3} & \textbf{88.6} \\
    DeepSeek R1 & 72.6 & 60.4 & 65.7 & 83.7 & 90.2 & 86.8 \\
    \bottomrule
    \end{tabular}
    }
\caption{Performance of three models on our dataset in a single run. \textbf{Llama}: 2-step chain-of-thought (CoT-2S) using Llama-70B. \textbf{GPT-4o}: CoT-2S using GPT-4o's API. \textbf{DeepSeek R1}: CoT-2S using distilled DeepSeek-R1-Distill-Llama-70B.
}
\label{table:deepseek_table}
\end{table}

\begin{table}[t]
\centering
\resizebox{0.99\linewidth}{!}{
\begin{tabular}{cccc}
\toprule
 & \textbf{Llama-8B} & \textbf{Llama-70B} & \textbf{DeepSeek-R1} \\ 
\midrule
Running Time (hrs) & 13 & 40 & 210 \\ 
\bottomrule
\end{tabular}}
\caption{Inference time comparison on our 6,000 sentences with different models: Llama-3.1-8B, Llama-3.1-70B, DeepSeek-R1-Distill-Llama-70B, all using the CoT-2S setting. The 8B model takes one NVIDIA A100 GPU and 70B model takes two NVIDIA A100 GPUs.}
\label{tab:runtime}
\end{table}

The results show that DeepSeek achieves noticeably better performance than Llama, with its F1-score being 3.4 points higher on metonymic examples. Llama models exhibit a high recall for metonymic sentences, as evident from Table \ref{table:main_table}. DeepSeek's results align more closely with GPT-4o, achieving higher precision but lower recall, indicating that it is more conservative at predicting metonymy. However, it is also worth mentioning that DeepSeek-R1-Distill-Llama-70B model takes around 5x longer time than Llama 70B to make predictions for all 6,000 sentences in our dataset, as shown in Table~\ref{tab:runtime}. For all models, we set the same maximum output lengths.

\section{Metonymy in Single Noun vs. Noun Phrase}
\label{app:bothmet}
Our prompts are designed to detect metonymy by identifying the transition on semantic categories associated with a  single noun in the sentence. One limitation of this concept is that sometimes the category change occurs in the noun phrases, rather than the single noun itself. This is particularly common in the \textsc{producer} category. In the sentence ``\textit{The \textbf{author} William Shakespeare is read all over the world}'', one can argue that the transition lies predominantly in the noun phrase ``\textit{author William Shakespeare}'', rather than just the noun ``\textit{author}''. 

Sentence (3) and (4) in Table \ref{tab:error_case} are examples where the noun is used metonymically within the noun phrase, but literally across the sentence. In sentence (3), the noun \textit{scientist} in the phrase ``\textit{most cited European social scientist}'' refers to the work of the scientist being cited. But in a broader context, scientist refers to the person themselves. Similarly, in the phrase ``\textit{Yiddish-speaking metropolis}'' in sentence (4), \textit{metropolis} refers to the people of the metropolis in this context, however it also refers to the location across the entire sentence. This highlights the diversity and challenges that our dataset offered, leaving room for future research to address these various linguistic instances. 

\section{Target Word Miscategorization}

The chain-of-thought prompt used in our CoT-2S model relies on the accuracy of the categorization performed in the first step. While the categorization is generally accurate, errors are more frequent for target words in the \textsc{possessed} and \textsc{causer} category. This is because these categories do not correspond to a well-defined semantic class, making them harder to categorize than the other four. Table \ref{tab:3col-example} presents the outputs of Llama-70B and GPT-4o using our CoT-2S architecture. 

\label{app:prompts}
\begingroup
\renewcommand\baselinestretch{0.98}
\begin{table*}[h]
    \NiceMatrixOptions {
    custom-line = {
       command = dashedmidrule ,
       tikz = { dashed } ,
       total-width = \pgflinewidth + \aboverulesep + \belowrulesep ,
     } }
    \centering
    \footnotesize
    \resizebox{\textwidth}{!}{
    \begin{NiceTabular}{|p{0.95\linewidth}|c|l}
    \toprule
    \multicolumn{1}{c|}{\bf Prompt} &
    \multicolumn{1}{c}{\bf Category} \\
    \midrule
    Metonymy is a figure of speech that substitutes the name of one thing for that of another of which it is an attribute or with which it is associated (such as ``bottle'' in ``liquid in the bottle''). You will be given a sentence and a target word. The category of the target word is a \textsc{container}. Your task is to determine if the target word in the sentence is used in a metonymic sense or literal sense. Think in the following steps: 
    
    1) Is the target word still used in the sense of a \textsc{container}? If yes, proceed to the next step. If not, then the sentence is not metonymic. In that case, do not perform the next steps, go to step 7 and say that it is \textsc{literal}.
    
    2) Determine if there is a shift of the semantic category, i.e., whether the true semantic meaning of that word still belongs to that category. 
    
    3) If there is a shift, has the semantic meaning of the word shifted from \textsc{container} to \textsc{content} or \textsc{process}? 
    
    4) Does the target word refer to another entity? 
    
    5) If the change in semantic meaning is from \textsc{container} to \textsc{content}, is the \textsc{content} already present in the sentence? If yes, it is not metonymic. So go to step 7 and say it is \textsc{literal}. 
    
    6) Use this comparison to determine if there is a metonymy in the sentence. 
    
    7) In the last step, give a one-word answer saying ``Final answer: \textsc{metonymic}'' or ``Final answer: \textsc{literal}''. & \textsc{container} \\
    \midrule
    Metonymy is a figure of speech that substitutes the name of one thing for that of another of which it is an attribute or with which it is associated (such as ``orchestra'' in ``music produced by the orchestra''). You will be given a sentence and a target word. The category of the target word is a \textsc{producer}. You need to identify if the target word in the sentence is used in a metonymic sense or literal sense. Think in the following steps: 
    
    1) Determine if there is a shift of the semantic category, i.e., whether the true semantic meaning of that word is still a \textsc{producer}. 
    
    2) If there is a shift, has the semantic meaning of the word shifted from \textsc{producer} to \textsc{product}? 
    
    3) Does the word refer to the producer itself? If so, then the sentence is \textsc{literal}. 
    
    4) Does the word refer to the product of the \textsc{producer}? In that case, the sentence is metonymic. 
    
    5) Use this comparison to determine if there is a metonymy in the sentence. 
    
    6) In the last step, give a one-word answer saying ``Final answer: \textsc{metonymic}'' or ``Final answer: \textsc{literal}''. & \textsc{producer} \\
    \midrule
    Metonymy is a figure of speech that substitutes the name of one thing for that of another of which it is an attribute or with which it is associated (such as ``magazine'' in ``editor or author producing the magazine''). You will be given a sentence and a target word. The category of the target word is a \textsc{product}. You need to identify if the target word in the sentence is used in a metonymic sense or literal sense. Think in the following steps: 
    
    1) Has the semantic meaning of the word shifted from \textsc{product} to \textsc{producer} or \textsc{organization}? 
    
    2) Check if the word refers to the producer of the product in any part of the sentence, instead of the physical product. (Example- ``the magazine criticized the man'', where ``magazine'' does not refer to the physical magazine but the producer of the magazine. It is metonymic in such a case). 
    
    3) Check if the word refers to the organization that creates the product in any part of the sentence, instead of the physical product. (Example- ``She worked for the Sun magazine'' where ``magazine'' refers to the company that produces the magazine and not the physical magazine). It is metonymic in such a case. 
    
    4) Determine whether the word is used in a metonymic sense using the above-mentioned steps. 
    
    5) In the last step, give a one-word answer saying ``Final answer: \textsc{metonymic}'' or ``Final answer: \textsc{literal}''. & \textsc{product} \\
    \midrule
    Metonymy is a figure of speech that substitutes the name of one thing for that of another of which it is an attribute or with which it is associated (such as ``city'' in ``residents of the city''). You will be given a sentence and a target word. The category of the target word is a \textsc{location}. You need to identify if the target word in the sentence is used in a metonymic sense or literal sense. Think in the following steps: 
    
    1) Determine if there is a shift of the semantic category, i.e., whether the true semantic meaning of that word has shifted from \textsc{location}. 
    
    2) Check if the target word (\textsc{location}) indicates the \textsc{people} of the location (Example - ``The city/island/pub hosted the event'' where ``city/pub/island'' indicates the \textsc{people} of that location hosting the event, and not the physical location). If so, then it is metonymic. 
    
    3) Check if the target word (\textsc{location}) refers to the \textsc{activity} (Example - ``He devotedly followed the church'' where ``church'' represents the beliefs of the church and not the physical church). If so, then it is metonymic. 
    
    4) Check if the target word (\textsc{location}) refers to an \textsc{institution} (Example; ``the department was notified'' where ``department'' refers to the institution and not the physical location). If so, then it is metonymic. 
    
    5) Use this comparison to determine if there is a metonymy in the sentence. 
    
    6) In the last step, give a one-word answer saying ``Final answer: \textsc{metonymic}'' or ``Final answer: \textsc{literal}'' & \textsc{location} \\
    \midrule
    Metonymy is a figure of speech that substitutes the name of one thing for that of another of which it is an attribute or with which it is associated (such as ``crown'' in ``the monarchy''). You will be given a sentence and a target word. You need to identify if the target word in the sentence is used in a metonymic sense or literal sense. Think in the following steps: 
    
    1) Is there a shift in the semantic meaning of the target word when it is used in the sentence? 
    
    2) If the meaning has shifted, does it refer to a different semantic category? If the semantic meaning has shifted to a category, then the word is used in a metonymic sense. Otherwise, the word is used in a literal sense. 
    
    3) Does the target word refer to something else of which it is an attribute of? 
    
    4) Use the comparison to determine if there is metonymy in the sentence? 
    
    5) In the last step, give a one-word answer saying ``Final answer: \textsc{metonymic}'' or ``Final answer: \textsc{literal}''  & \textsc{general} \\
    \bottomrule
    
    \end{NiceTabular}
    }
    \caption{Prompts used in our models. In CoT model, only the \textsc{general} prompt is used, while all categories are used in the CoT-2S model.}
    \label{tab:all_prompts}
\end{table*}
\endgroup

\begingroup
\renewcommand\baselinestretch{0.9}

\NiceMatrixOptions{
  custom-line = {
    command = dashedmidrule,
    tikz = { dashed },
    total-width = \pgflinewidth + \aboverulesep + \belowrulesep,
  }
}

\begingroup
\renewcommand{\arraystretch}{2}
\onecolumn
{\small
\begin{longtable}{|p{0.20\linewidth}|p{0.35\linewidth}|p{0.35\linewidth}|}
\caption{Examples comparing Llama-70B and GPT-4o with CoT-2S outputs.}\label{tab:3col-example}\\

\hline
\bf Example Information & \bf Llama-70B Output & \bf GPT-4o Output \\
\hline
\endfirsthead

\multicolumn{3}{l}{\textit{(Continued from previous page)}}\\
\hline
\bf Example Information & \bf Llama-CoT-2S Output & \bf GPT-4o Output \\
\hline
\endhead

\hline
\multicolumn{3}{r}{\textit{(Continued on next page)}}\\
\hline
\endfoot

\hline
\endlastfoot


\tb{Sentence}: He also enjoyed a \textbf{glass} of whisky or wine while relaxing. \newline
\tb{Target Word}: glass \newline
\tb{Category}: \textsc{container} \newline
\tb{Label}: 

\tr{\textsc{\textbf{non-metonymic}}}

&

\textsc{container} \newline
To determine if the target word ``glass'' is used in a metonymic or literal sense, let's follow the steps:

1. The target word ``glass'' is still used in the sense of a \textsc{container}. Proceed to the next step.

2. There is a shift of the semantic category. The true semantic meaning of ``glass'' in this context is not just the container itself, but also what it contains.

3. The semantic meaning of the word has shifted from \textsc{container} to \textsc{content}.

4. The target word ``glass'' refers to another entity, which is the whisky or wine it contains.

5. The \textsc{content} (whisky or wine) is already present in the sentence, but it is not explicitly stated that the glass contains whisky or wine. However, the context implies that the glass is not empty.

6. Based on the comparison, there is a metonymy in the sentence. The word ``glass'' is used to refer to the whisky or wine it contains, rather than just the container itself.

7. Final answer: \tg{\textsc{\textbf{metonymic}}}

&

\textsc{container} \newline
To determine if the target word ``glass'' is used in a metonymic or literal sense, let's follow the steps systematically:

1) Is the target word still used in the sense of a \textsc{container}? - ``Glass'' typically refers to a container, but in this sentence, ``a glass of whisky or wine'' refers to the liquid inside rather than the container itself. Since ``glass'' is not being used as a container, we move to the next step.

2) Determine if there is a shift of the semantic category. - Yes, ``glass'' has shifted from being a \textsc{container} to referring to its \textsc{content} (whisky or wine)

3) Has the semantic meaning shifted from \textsc{container} to \textsc{content} or \textsc{process}? - Yes, it has shifted to \textsc{content}.

4) Does the target word refer to another entity? - Yes, ``glass'' refers to the liquid inside it.

5) If the change is from \textsc{container} to \textsc{content}, is the \textsc{content} already present in the sentence? - Yes, the \textsc{content} (``whisky or wine'') is explicitly mentioned in the sentence.

6) Use this comparison to determine if there is metonymy. - Since the \textsc{content} is already mentioned, this is a literal use of ``glass'' rather than a metonymic shift.

Final Step: Final answer: \tr{\textsc{\textbf{literal}}} \\

\hline


\tb{Sentence}: His publications have been cited several ten thousand times, which makes him one of the currently most cited European social \textbf{scientist}. \newline
\tb{Target Word}: scientist \newline
\tb{Category}: \textsc{producer} \newline
\tb{Label}: \tg{\textsc{\textbf{metonymic}}}

&
       
\textsc{producer} \newline
To determine if the target word ``scientist'' is used in a metonymic or literal sense, let's follow the steps:

1. Determine if there is a shift of the semantic category: The true semantic meaning of ``scientist'' is a \textsc{producer}, referring to a person who conducts scientific research.

2. If there is a shift, has the semantic meaning of the word shifted from \textsc{producer} to \textsc{product}? In this sentence, there is no apparent shift in the semantic category. The word ``scientist'' still refers to a person.

3. Does the word refer to the producer themselves? Yes, in this sentence, ``scientist'' refers to the person who conducts scientific research, which is the producer.

4. Does the word refer to the product of the producer? No, the sentence does not refer to the product of the scientist's research, but rather the scientist themselves.

5. Use this comparison to determine if there is a metonymy in the sentence: Since the word ``scientist'' refers to the producer themselves and not their product, there is no metonymy.

6. Final answer: \tr{\textsc{\textbf{literal}}}

&

\textsc{producer} \newline
To determine if the target word ``scientist'' is used in a metonymic or literal sense, let's analyze the sentence using the given steps:

1) Determine if there is a shift of the semantic category (i.e., whether the true semantic meaning of the word is still a \textsc{producer}). - The word ``scientist'' refers to a person who produces research and publications. In this sentence, ``scientist'' refers to the person himself and not the product (publications). Since there is no category shift, we proceed to step 3.

2) If there is a shift, has the semantic meaning of the word shifted from \textsc{producer} to \textsc{product}? - Skip, as no shift has occurred

3) Does the word refer to the producer itself? - Yes, ``scientist'' refers to the person and not their work.

4) Does the word refer to the product of the producer? - No, the word does not refer to the product.

5) Use this comparison to determine if there is a metonymy in the sentence - Since ``scientist''  directly refers to the person and not their publications, the usage is literal..

Final Step: Final answer: \tr{\textsc{\textbf{literal}}} \\

\hline

\tb{Sentence}: The novel is set in Sitka, which it depicts as a large, Yiddish-speaking \textbf{metropolis}. \newline
\tb{Target Word}: metropolis \newline
\tb{Category}: \textsc{location} \newline
\tb{Label}: \tg{\textsc{\textbf{metonymic}}}

&

\textsc{location} \newline
To determine if the target word ``metropolis'' is used in a metonymic sense or literal sense, let's follow the steps:

1. Determine if there is a shift of the semantic category: The word ``metropolis'' is a \textsc{location}, and it refers to a large city, which is a type of location. There is no apparent shift in the semantic category.

2. Check if the target word indicates the PEOPLE of the location: In this sentence, ``metropolis'' does not indicate the people of the location. It describes the location itself as a large, Yiddish-speaking city.

3. Check if the target word refers to the \textsc{activity}: ``metropolis'' does not refer to any activity in this sentence.

4. Check if the target word refers to an \textsc{institution}: ``metropolis'' does not refer to an institution in this sentence.

5. Comparison: Based on the above analysis, ``metropolis'' is used to describe the physical location and its characteristics, without any shift in the semantic category or indication of people, activity, or institution.

6. Final answer: \tr{\textsc{\textbf{literal}}}

&

\textsc{location} \newline
To determine if the target word ``metropolis'' is used in a metonymic or literal sense, let's follow the steps:

1) Determine if there is a shift of the semantic category from \textsc{location}.- The word ``metropolis'' is a type of \textsc{location}, and in this sentence, it refers to a city. There is no apparent shift in the semantic category.

2) Check if the target word \textsc{location} indicates the \textsc{people} of the location. - In this sentence, ``metropolis'' refers to the city itself, not the people living in it.

3) Check if the target word \textsc{location} refers to the \textsc{activity}. - There is no indication that ``metropolis'' represents an activity in this sentence.

4) Check if the target word \textsc{location} refers to an \textsc{institution}. - ``Metropolis'' does not refer to an institution in this context.

5) Use this comparison to determine if there is a metonymy in the sentence. - Based on the analysis, the word ``metropolis'' is used to describe the city of Sitka, focusing on its characteristics as a large, Yiddish-speaking city. There is no metonymic shift in meaning.

Final step: Final answer: \tr{\textsc{\textbf{literal}}} \\

\hline


\tb{Sentence}: They will be flown into the stadium with helicopters when a full \textbf{stadium} is cheering for them when they arrive. \newline
\tb{Target Word}: stadium \newline
\tb{Category}: \textsc{location} \newline
\tb{Label}: \tg{\textsc{\textbf{metonymic}}}

&

\textsc{location} \newline
To determine if the target word ``stadium'' is used in a metonymic or literal sense, let's analyze the sentence step by step.

1. The semantic category of the word ``stadium'' is indeed \textsc{location}.

2. In the given sentence, ``a full stadium is cheering for them'' indicates the \textsc{people} in the stadium, not the physical location itself. This implies a shift in the semantic meaning from \textsc{location} to the \textsc{people} in that location.

3. There is no indication that ``stadium'' refers to an \textsc{activity} in this context.

4. There is also no indication that ``stadium'' refers to an \textsc{institution} in this context.

5. Based on the analysis, the word ``stadium'' is used to represent the \textsc{people} in the stadium, which is a metonymic sense.

Final answer: \tg{\textsc{\textbf{metonymic}}}

&

\textsc{location} \newline
To determine if the target word ``stadium'' in the sentence is used in a metonymic sense or a literal sense, let's follow the outlined steps:

1) Determine if there is a shift of the semantic category from \textsc{location}. - The term ``stadium'' generally refers to a large structure or venue designed for sports or concerts.

2) Check if the target word \textsc{location} indicates the \textsc{people} of the location. - The sentence mentions ``stadium'' in the context of hosting concerts. It does not seem to refer to the people associated with the stadium but rather the venue itself.

3) Check if the target word \textsc{location} refers to the \textsc{activity}. - ``stadium'' here does not refer to an activity. Instead, it is used as the place where an activity (concert) occurred.

4) Check if the target word \textsc{location} refers to an \textsc{institution}. - ``stadium'' in this context does not seem to refer to an institution. It refers to the physical venue where the concerts took place.

5) Use this comparison to determine if there is a metonymy in the sentence. - As per the above analysis, ``stadium'' is used to describe the physical location where the concert occurred. There is no shift from the semantic category of \textsc{location} to another category such as people, activity, or institution.

Final step: Final answer: \tr{\textsc{\textbf{literal}}} \\

\hline


\tb{Sentence}: As he began his questioning of the witnesses, the Clodian \textbf{crowd} drowned out his voice and surrounded him. \newline
\tb{Target Word}: crowd \newline
\tb{Category}: \textsc{causer} \newline
\tb{Label}: \tg{\textsc{\textbf{metonymic}}}

&

\textsc{container} \newline
To determine if the target word ``crowd''in the sentence is used in a metonymic sense or literal sense, let's break it down:

1. Is the target word ``crowd'' still used in the sense of a \textsc{container}? No, a crowd is a group of people, not a container.

Since the target word is not used as a container, we can conclude that it's not metonymic in the context of a container. However, we'll follow the steps for clarity.

2. Determine if there is a shift of the semantic category: Yes, the word ``crowd'' refers to a group of people, not a container.

3. If there is a shift, has the semantic meaning of the word shifted from \textsc{container} to \textsc{content} or \textsc{process}? Not applicable, as ``crowd'' is not a container.

4. Does the target word refer to another entity? Yes, the target word ``crowd'' refers to a group of people.

5. Not applicable, as the content is not present in the sentence in relation to a container.

6. Not applicable, as the target word is not a container.

7. Final answer: \tr{\textsc{\textbf{literal}}}
&
\textsc{general} \newline
1) Shift in the semantic meaning: The word ``crowd'' usually refers to a group of people gathered together. In this case, the meaning remains the same, referring to the group of people.

2) Meaning shift to a different category? - The meaning of ``crowd'' does not shift to a different category. It still refers to people and not to an abstract concept or attribute of an institution.

3) Does it refer to something else of which it is an attribute? - No, ``crowd'' in this case is being used to refer to a group of people, not to something abstract or symbolic of an institution.

4) Metonymy comparison - There is no indication that the word ``crowd'' is being used as a substitute for something associated with an institution. It refers directly to the people present.

Final answer: \tr{\textsc{\textbf{literal}}} \\

\hline


\tb{Sentence}: Holland first heard the Native American \textbf{flute} at a concert Webster University near St. Louis in 1994. \newline
\tb{Target Word}: flute \newline
\tb{Category}: \textsc{causer} \newline
\tb{Label}: \tg{\textsc{\textbf{metonymic}}}

&

\textsc{product} \newline
To determine whether the target word ``flute'' is used in a metonymic or literal sense, let's analyze the sentence:

1. The semantic meaning of 'flute' has not shifted from \textsc{product} to \textsc{producer} or \textsc{organization} in this context.

2. The word ``flute'' does not refer to the producer of the flute in the given sentence. It refers to the musical instrument itself.

3. The word ``flute'' does not refer to the organization that creates the flute in the given sentence. It refers to the musical instrument itself.

4. Based on the analysis, ``flute'' is used to refer to the musical instrument, not the producer or organization.

Final answer: \tr{\textsc{\textbf{literal}}}

&

\textsc{product} \newline
To determine if the target word ``flute'' is used in a metonymic sense or literal sense, let's analyze the sentence using the given steps:

1) Has the semantic meaning of the word shifted from \textsc{product} to \textsc{producer} or \textsc{organization}? - The semantic meaning of the word ``flute'' has not shifted from \textsc{product} to \textsc{producer} or \textsc{organization} in this context.

2) Check if the word refers to the producer of the product in any part of the sentence, instead of the physical product. - The word ``flute'' does not refer to the producer of the flute in any part of the sentence. Instead, it refers to the musical instrument itself.

3) Check if the word refers to the organization that creates the product in any part of the sentence, instead of the physical product. - The word ``flute'' does not refer to the organization that creates the flute in any part of the sentence. It is specifically referring to the Native American flute as a musical instrument.

4) Determine whether the word is used in a metonymic sense using the above-mentioned steps. - Based on the analysis, the word ``flute'' is used to refer to the actual musical instrument, not the producer or organization.

Final step: Final answer: \tr{\textsc{\textbf{literal}}} \\

\hline


\tb{Sentence}: He nearly spun out with 20 laps to go but saved the \textbf{truck}, later inheriting the lead from Stewart Friesen after he ran out of fuel. \newline
\tb{Target Word}: truck \newline
\tb{Category}: \textsc{possessed} \newline
\tb{Label}: 

\tr{\textsc{\textbf{non-metonymic}}}

&

\textsc{container} \newline
To determine if the target word ``truck'' in the sentence is used in a metonymic sense or literal sense, we'll follow the steps:

1. The target word ``truck'' is still used in the sense of a \textsc{container} (vehicle), so we proceed to the next step.

2. There is a shift of the semantic category. The true semantic meaning of ``truck'' in this context is not just a vehicle, but the entity that is racing or participating in the event.

3. The semantic meaning of the word has shifted from \textsc{container} (vehicle) to \textsc{content} (the entity participating in the event, which includes the driver).

4. The target word ``truck'' refers to another entity, which is the driver or the team participating in the event.

5. The \textsc{content} (the entity participating in the event) is not explicitly present in the sentence, but it is implied.

6. Comparing the original meaning of ``truck'' as a \textsc{container} (vehicle) and its shifted meaning to \textsc{content} (the entity participating in the event), we can see that there is a metonymic relationship.

7. Final answer: \tg{\textsc{\textbf{metonymic}}}

&

\textsc{container} \newline
To determine if the sentence is metonymic, let's follow the steps:

1) Is the target word still used in the sense of a \textsc{container}? - The word ``truck'' refers to a physical vehicle, which can act as a container by holding cargo or passengers. Since the sentence mentions ``saved the truck,'' it still refers to the truck as a vehicle rather than something else. So, we proceed to the next step.

2) Determine if there is a shift of the semantic category, i.e., whether the true semantic meaning of that word still belongs to that category. - ``truck'' in this sentence refers to the physical vehicle, which is still a \textsc{container}. There is no evident shift in meaning.

3) If there is a shift, has the semantic meaning of the word shifted from \textsc{container} to \textsc{content} or \textsc{process}? - No, ``truck'' is still being used as a vehicle, not as its contents (such as cargo) or as a process.

4) Does the target word refer to another entity? - No, ``truck'' still refers to the actual vehicle.

5) If the change in semantic meaning is from \textsc{container} to \textsc{content}, is the \textsc{content} already present in the sentence? - Since no such shift has occurred, this step is not applicable.

6) Use this comparison to determine if there is a metonymy in the sentence - There is no metonymic shift in meaning. The word ``truck'' is used literally to refer to the vehicle. Final answer: The word ``truck'' is used in its literal sense as a vehicle.

Final answer: \tr{\textsc{\textbf{literal}}} \\

\hline


\tb{Sentence}: 1942, known as 48 Hours in the USA, in which she is shown wielding a \textbf{rifle} to defend a house from German paratroopers. \newline
\tb{Target Word}: rifle \newline
\tb{Category}: \textsc{possessed} \newline
\tb{Label}: 

\tr{\textsc{\textbf{non-metonymic}}}

&

\textsc{producer} \newline
To determine if the sentence contains a metonymy, let's go through the steps:

1. Determine if there is a shift of the semantic category: In this case, the semantic category of the word ``rifle'' is still a \textsc{product}, specifically a type of firearm.

2. If there is a shift, has the semantic meaning of the word shifted from \textsc{producer} to \textsc{product}? Yes, the word ``rifle'' refers to the product itself, not the producer.

3. Does the word refer to the producer themself? No, the word ``rifle'' does not refer to the producer.

4. Does the word refer to the product of the producer? Yes, the word ``rifle'' refers to the product of the producer, in this case, the firearm used by the character.

5. Use this comparison to determine if there is a metonymy in the sentence: Since the word ``rifle'' refers to the product of the producer, and not the producer themselves, this is a case of metonymy.

6. Final answer: \tg{\textsc{\textbf{metonymic}}}

&

\textsc{general} \newline
1) The target word ``rifle'' generally refers to a firearm used for shooting. In the given sentence, it is used in the phrase ``wielding a rifle to defend a house from German paratroopers.''

2) Here, ``rifle'' retains its core meaning as a firearm.
The semantic meaning does not shift to a different category. It still refers to a physical object, a weapon used in combat.

3) ``Rifle'' does not refer to something else of which it is an attribute. It is not used to represent an institution, a group of people, or an abstract concept.

4) Since there is no shift in meaning, no different semantic category, and no attribute-based substitution, there is no metonymy in this sentence.

Final answer: \tr{\textsc{\textbf{literal}}} \\

\end{longtable}
}
\endgroup

\end{document}